\documentclass[conference]{IEEEtran}
\usepackage{blindtext, graphicx}
\usepackage{amsmath} 
\usepackage{float}

\newcommand{\conj}{\mathrm{conj}}
\newcommand{\mean}{\mathrm{mean}}

\newcommand{\betP}{\mathrm{betP}}

\begin{document}
%
\title{An automatic water detection approach based on Dempster-Shafer theory for multi-spectral images}

\author{\IEEEauthorblockN{Na Li}
\IEEEauthorblockA{IRISA - University of Rennes 1 - Total\\
France\\
Email: li.na@irisa.fr}
\and
\IEEEauthorblockN{Arnaud Martin}
\IEEEauthorblockA{IRISA - University of Rennes 1\\
France\\
Email: arnaud.martin@irisa.fr}
\and
\IEEEauthorblockN{R\'emi Estival}
\IEEEauthorblockA{Total, Innovative acquisition\\ 
France\\
Email: remi.estival@total.com}}


%


\maketitle

\begin{abstract}
Detection of surface water in natural environment via multi-spectral imagery has been widely utilized in many fields, such land cover identification. However, due to the similarity of the spectra of water bodies, built-up areas, approaches based on high-resolution satellites sometimes confuse these features. A popular direction to detect water is spectral index, often requiring the ground truth to find appropriate thresholds manually. As for traditional machine learning methods, they identify water merely via differences of spectra of various land covers, without taking specific properties of spectral reflection into account. In this paper, we propose an automatic approach to detect water bodies based on Dempster-Shafer theory, combining supervised learning with specific property of water in spectral band in a fully unsupervised context. The benefits of our approach are twofold. On the one hand, it performs well in mapping principle water bodies, including little streams and branches. On the other hand, it labels all objects usually confused with water as `ignorance', including half-dry watery areas, built-up areas and semi-transparent clouds and shadows. `Ignorance' indicates not only limitations of the spectral properties of water and
supervised learning itself but insufficiency of information from multi-spectral bands as well, providing valuable information for further land cover classification.

\end{abstract}

\begin{IEEEkeywords}
Surface water detection, multi-spectral image, Dempster-Shafer theory.
\end{IEEEkeywords}

%
\IEEEpeerreviewmaketitle

\section{Introduction}
\label{int}
Detection of surface water through multi-spectral data is an important topic in remote sensing. In recent years, various approaches have been developed to detect water on the ground, especially for small water bodies such as streams. Spectral index is one of the most popular directions in water detection, taking advantage of differences of land covers in spectral reflection~\cite{index}. 
Many indexes could be combined together to extract features of water bodies and to detect changes~\cite{rs6054173}. The normalized difference vegetation index (NDVI), a satellite-derived index from the Near-Infrared (NIR) and Red bands is the most widely used index for mapping land covers, often indicating water with negative values~\cite{rs6010580}.
However, no universal thresholds of NDVI exsit to identify land covers and plenty of ground truth is required to verify various thresholds manually as well. The normalized difference water index (NDWI) derived from Near-Infrared and Short Wave Infrared (SWIR) bands is also known to be strongly related to water content yet preliminary experiments have proved that NDWI may confuse water and built-up areas due to their similar characteristics~\cite{ndwi}. Additionally, some specific features from spectral bands also provide useful information to identify land covers. As NIR energy tends to be absorbed strongly by water than by others ({\em e.g.} vegetation, built-up areas, {\em etc}), this specific property is usually utilized to identify water. Based on the property, a spectral model could be applied to automatically find a threshold in NIR channel to distinguish water~\cite{rs6054174}. 

On the other hand, machine learning methods are also frequently applied in identification of water bodies. Given appropriate and adequate training samples, supervised learning shows a good performance, for instance, support vector machine (SVM), neural network, and decision tree~\cite{dt}. However, unlike to other land covers, surface water could produce diverse spectral response from large rivers to small streams, which makes training samples be hard to find adequately. Unsupervised learning is difficult to apply directly on water detection because it separates data in clusters rather than identifies the specific land cover types. Moreover, since ground truth is unavailable in some cases, it is of great importance to detect water automatically in a fully unsupervised context.

As spectral signatures do not always provide enough information in classification decisions, multi-sensor data fusion has become another promising aspect in land cover classification. Plenty of approaches have been proposed to combine information from various sensors. In general, fusion of multi-sensor data can be classified into three different levels: the pixel level, the feature level and the decision level.

Fusion in the pixel level consists in considering different original data from multiple sensors as the data from one signal source with single resolution, making data more informative than an individual source~\cite{Solanky2016}. In the feature level, several features ({\em e.g.} edges, lines and texture information) are extracted from different data sources so that they can be combined into one or more features maps, rendering more information than original data. Fusion in this level is of great importance when numerous spectral bands are available, avoiding to analyze each band separately.
Decision level fusion combines the classification results from multiple sensors to generate a final decision. Plenty of theories have been proposed and developed in this level, for instance, probabilistic methods, theory of possibilities and theory of belief functions~\cite{4683343,823918}. 

The theory of belief functions, also called Dempster-Shafer theory, has been wildly used in multi-sensor data fusion as a method in the decision level. It performs well in merging classification results from multiple sources, owing to the measurement of uncertainty and imprecision~\cite{Martin04a}. For instance, an evidential model is proposed to deal with the statistical segmentation of multi-sensor images, taking into account contextual information via Markovian fields~\cite{942557}. An incorporation of Landsat TM imagery, altitude and slope data through evidential reasoning improved classification accuracy thanks to uncertainty introduced in the classification system~\cite{Li2011}. A multidimensional evidential reasoning (MDER) approach was proposed to estimate change detection from the fusion of heterogeneous remote sensing images~\cite{6495471}.  Dempster-Shafer theory has also been used to relax Bayesian decisions given by a Markovian classification algorithm (ICM), which shows satisfying performances in the classification of very noisy remote sensing images~\cite{997824}. In addition, for the combination of multi-scale data, an algorithm based on Dempster-Shafer theory allows to model the mixed feature of the low spatial resolution pixels and the class confusion, taking compound hypotheses into consideration~\cite{Hegarat-Mascle:2003:MDF:1273240.1273241}. In general, Dempster-Shafer theory is usually utilized to fuse classification results from supervised learning, while it is also promising to apply it in a fully unsupervised context. In~\cite{602544}, the authors proposed a fusion between two unsupervised learning methods performing well in separating land covers.

In the Dempster-Shafer framework, each sensor may have different reliability and importance in application. Therefore, it is important to weaken results from sensors so that some unreliable perspectives would not be overestimated during fusion. An estimation method of discounting coefficient for multisensors is proposed by using
dissimilarity measure~\cite{Liu2011133}. When applying Dempster-Shafer framework on land covers classification, spatial information also plays an important role in satellite image processing, contributing to decreasing noisy pixels and adding texture information as well. The neighborhood relationship is often used to provide a more accurate modeling of the information~\cite{5595234}.


To detect water in a fully unsupervised learning, we propose a new automatic approach
based on the fusion of a spectral model and supervised learning. The training samples of supervised learning is generated from classification results of the spectral model. Thus, Dempster-Shafer theory is used in the situation where two sources are dependent~\cite{Chebbah2012}. Discounting coefficients are used as penalty for results of the spectral model, considering results from supervised learning as the ground truth. In addition, in order to add spatial information during fusion, we propose a coefficient in mass function based on labels of neighbors in a rectangular window. 

This paper is organized as follows: In section~\ref{DS}, some bases of Dempster-Shafer theory are recalled, followed by the explanation of two water detection models explained in section~\ref{waterDM}. Sections~\ref{BBAModel} and~\ref{Fusion} present the principle methodology, including construction of mass function for the two models introduced in section~\ref{waterDM} and presentation of details in fusion and decision. Then, section~\ref{expe} illustrates results of experiments. Conclusions are drawn in section~\ref{conclusion}.

\section{Dempster-Shafer theory}
\label{DS}

As a generalization of traditional probability, Dempster-Shafer theory~\cite{dempster1967,shafer1976mathematical} allows to distribute support for proposition not only to a single proposition itself but also to the union of propositions that include it. One of the greatest advantages of Dempster-Shafer theory is that it allows to take into consideration uncertainty and imprecision at the same time {\em via} two functions: belief and plausibility, derived from mass function. The mass function is defined on all the subsets of the frame of discernment $\Omega=\{\omega_1, \ldots, \omega_n$\}, and assigns belief degree to all the elements in the power set of discernment, noted as $2^\Omega$. 

The mass function of the null proposition $\emptyset$ is usually set to zero but it is also possible to be a positive value. The sum of the masses of all the propositions is one:
\begin{equation}
\sum_{A\subseteq\Omega}m(A) = 1
\end{equation}
$m(\emptyset)=0$ refers to a closed world hypothesis, in which the discernment $\Omega$ contains all the possible situations in reality. On the contrary, if $m(\emptyset)$ is superior to $0$, this corresponds to an open world hypothesis, where unknown situation outside of $\Omega$ can be considered. $m(A)$ can be considered as a degree of evidence supporting the claim that a specific element of $\Omega$ belongs to the set $A$, yet not to any subset of $A$~\cite{Klir:1987:FSU:30395}.

The belief in a proposition is the sum of masses of all propositions contained in it, which can be interpreted as the total amount of justified support given to this proposition~\cite{DENOEUX199979}. The plausibility of a proposition is the sum of the masses of all propositions in which it is wholly or partly contained, which can be interpreted as the maximum amount of specific support that could be give to this proposition~\cite{Smets:1994:TBM:179151.179153}. The belief function $Bel(A)$ and plausibility function $Pl(A)$ are therefore defined by:
\begin{equation}
Bel(A) = \sum_{X\subseteq\ A}m(X)
\end{equation}
\begin{equation}
Pl(A) = \sum_{X\cap A\neq\emptyset}m(X)
\end{equation}
In order to combine independent sources, the main combination rule is the conjunctive rule given $\forall A\subseteq \Omega$, by:
\begin{equation}
\label{conjunctive}
m_\conj(A) = \sum\limits_{X_1 \cap \cdots \cap X_S = A} \prod_{s=1}^{S}m_s(X_s),
\end{equation}
where $s$ represents the different sources from $1$ to $S$. 
For dependent sources, an idempotent rule must be used such as the average rule of combination given $\forall A\subseteq \Omega$ by:
\begin{equation}
    m_{\mean} (A) = \dfrac{1}{S}\displaystyle\sum_{s=1}^{S}m_s(A)
\end{equation}
The cautious rule of Den{\oe}ux~\cite{Denoeux2006} is also available for dependent sources.

For the decision step, the pignistic probability~\cite{Smets:1990:CPP:647232.719592} is currently used because it offers a good compromise between the maximum of credibility and the maximum of plausibility. The basic idea of pignistic probability is to dissipate the mass values associated with focal elements to a specified focal element, which has been generalized in Dempster-Shafer framework, given by:
\begin{equation}
    \betP(A) = \displaystyle\sum_{B\in 2^{\Omega},B\cap A\neq \emptyset} \dfrac{1}{\lvert B \rvert} \dfrac{m(B)}{1-m(\emptyset)}
\end{equation}
where $\lvert B \rvert$ represents the cardinality of $B$.

However, the pignistic probability do not allow to decide on composite hypotheses. The proposed approach by Appriou~\cite{Appriou} is adapted to this kind of decision. The principle is to weight the decision function, such as the pignistic probability, by an utility function relying on the cardinality of the elements. For $L\in 2^\Omega$ is chosen as the label if:
\begin{equation}
\label{ap}
    L = \underset{X\in 2^\Omega}{\mathrm{argmax}} (m_d(X) \betP(X))
\end{equation}
where $m_d(X)$ is a mass defined by:
\begin{equation}
\label{md}
    m_d(X) = \frac{K_d\lambda_X}{\lvert X \rvert^r}
\end{equation}
$K_d$ is a normalization factor and $\lambda_X$ is applied to integrate the lack of knowledge about one of the elements of $2^\Omega$. $\lvert X \rvert$ stands for the cardinality of $X$. The value $r$ ranges from $0$ to $1$, allowing to choose a decision which varies from a total ignorance when $r$ is $0$ and a decision based on a singleton with $r$ is equal to $1$.

\section{Water detection models}
\label{waterDM}
In this section, we present two different models to identify water: the spectral model and the supervised model. Their results are combined through Dempster-Shafer theory.

The spectral model to identify water makes use of specific property of water in spectral bands. As water has the strongest absorption in NIR channel, its NIR reflection is able to show a great difference compared to other land covers. Based on this property, a threshold can be detected automatically to identify water and non-water. Water pixels correspond to inferior values to this threshold in NIR. On the contrary, non-water pixels reflect superior values to the threshold~\cite{rs6054174}. The method to find the threshold is explained as follows:
\begin{enumerate}[]
\item Calculate the histogram of NIR band
\item Find the two first local peaks in the NIR histogram
\item Use a five-degree polynomial function to approximate the part between the two local peaks
\item Find the minimal of the five-degree polynomial approximation and use its correspondent NIR value as the threshold of water.
\end {enumerate}

Although the threshold in NIR band allows water to be distinguished from other land covers, very thin clouds and shadows on vegetation could be confused with small and shallow water bodies sometimes. With merely information from NIR band, these confusing objects may be difficult to identify from each other while it is still meaningful to gather them as a group for further study. The confusing objects always have NIR values approaching to the threshold, while principle water pixels and other land covers with obvious distinction preserve some distance from the threshold. However, this distance is difficult to define. In this paper, we manage to tackle the problem from a new perspective.

The supervised model includes two steps: the learning step and the classification step. In the learning step, data associated with already known label is utilized to train parameters in the model. The classification step allows to predict labels of new data based on the learning function. However, the lack of learning data or the availability of inappropriate samples often leads to wrong classification and low accuracy. This problem is more pronounced in water detection since reference data is often chosen from satellite image, which results easily in lack of enough information of surface water. Large water bodies, such as rivers or lakes, are often easy to be detected, while smaller ones like streams tend to be confused with other land covers since not sufficient information is supported to represent their own spectral reflections. In this case, lacking information could increase imprecision and uncertainty of classification, causing unsatisfying results. In our study, on account of unavailability of the ground truth, a new approach is proposed to accomplish the detection of water in a fully unsupervised context. 

Considering original spectrum ({\em e.g.} Red, Green, Blue and Near-infrared(NIR)) is not sensitive enough to identify land covers, the supervised learning could achieve better performances in a transformed feature space. The feature space is composed of specific indexes extracted from the original bands. We utilize three indexes: NDVI, NDWI, the Red Edge Normalized Difference Water Index (RE\_NDWI)~\cite{rendwi} to construct the new feature space. As we have explained in section \ref{int}, NDVI is strongly related to vegetation and reflects difference of basic land covers as well. NDWI improves the separability of water from vegetation and soil while sometimes it is not able to distinguish water from built-up areas efficiently. In order to overcome this drawback, we use another index called RE\_NDWI to increase the separability of water from built-up areas. The three indexes are calculated as follows:
\begin{equation}
    NDVI = \frac{NIR-RED}{NIR+RED}
\end{equation}
\begin{equation}
    NDWI = \frac{NIR-GREEN}{NIR+GREEN}
\end{equation}
\begin{equation}
    RE\_NDWI = \frac{GREEN-RE}{GREEN+RE}
\end{equation}
where $RED$, $GREEN$ respectively stand for spectral reflectance measurements acquired in the visible red and green bands.
$NIR$, $RE$ individually represent reflectance from near-infrared and red-edge bands. Due to the lack of Shortwave Infrared (SWIR) band in our study, we calculate $NDWI$ through green and $NIR$ bands~\cite{rs6054174}. In our proposed approach, the supervised learning is conducted in the three dimensional space composed of $NDVI$, $NDWI$ and $RE\_NDWI$.

\section{Modelling of mass function}
\label{BBAModel}

As the spectral model is merely based on a threshold in NIR band, we define its mass function through the distance from each pixel to the threshold generated by the spectral model in NIR band. For a pixel labelled as `water', the closer to the threshold it is, the more uncertain it is to belong to `water' class. However, this does not mean it is more reliable to label the pixel as `non-water', for the reason that this pixel does not have a superior NIR value than the threshold. Therefore, mass value of a pixel to `non-water' class is set to zero on condition that NIR value of the pixel is inferior to the NIR threshold. This is the same for a pixel labelled as `non-water' class. A small yet non-zero mass value for `water' or `non-water' indicates that the pixel is indistinguishable due to the limitation of the spectral model, which is `ignorance'.

The discernment is defined as $\Omega=\{\omega_1, \omega_2\}$ where $\omega_1$ represents `water', $\omega_2$ `non-water'. The mass functions for pixel $x$ belonging to `water' class, `non-water' class are defined as follows:
$\forall x, n_x \leq t$
\begin{equation}
   m_1(\{\omega_1\})(x)= \frac{\alpha_{\omega_1}}{N}(1-e^{-\gamma_x\frac{(t-n_x)}{D_{\omega_1}}})
\end{equation}
\begin{equation}
    m_1(\{\omega_2\})(x)=0
\end{equation}
$\forall x, n_x >t$
\begin{equation}
    m_1(\{\omega_1\})(x) =0
\end{equation}
\begin{equation}
    m_1(\{\omega_2\})(x)=\frac{\alpha_{\omega_2}}{N}(1-e^{-\gamma_x\frac{(n_x-t)}{D_{\omega_2}}})
\end{equation}
For the discernement $\Omega$, the mass function is defined as follows:
\begin{equation}
   m_1(\{\Omega\})(x)=1-m_1(\{\omega_1\})(x)-m_1(\{\omega_2\})(x)
\end{equation} 
%
where $t$ is the threshold in NIR, $n_x$ is the NIR value of pixel $x$. $N$ is a normalization coefficient to make mass value range from 0 to 1, given by:
\begin{equation}
   N=1-e^{-1}
\end{equation}
The coefficients $\alpha_{\omega_1}$ and $\alpha_{\omega_2}$ are two individual discounting coefficients for `water' class and `non-water' class. The discounting coefficients are generated through the comparison between results from the spectral model and supervised learning, using confusion matrix. In the confusion matrix, results from supervised learning are considered as the ground truth, which will be explained in section \ref{Fusion}.

Now that the threshold is located around the lower end of NIR band, a great difference exists between the distance from the threshold to the pixels in its left side and its right side, requiring a normalized step for the two types of distances. The coefficients $D_{\omega_1}$ and $D_{\omega_2}$ are normalized coefficients for `water' and `non-water' pixels individually, defined through the largest distance from each end of NIR axis to the threshold. These coefficients are given by:
\begin{equation}
    D_{\omega_1} = t-n_{min}
\end{equation}
\begin{equation}
     D_{\omega_2} = n_{max}-t
\end{equation}
in which $n_{min}$ stands for the minimal value in NIR and $n_{max}$ represents the maximum value in NIR.

In addition, taking into consideration labels from neighbors as spatial information, mass function could be more effective to reflect belief degree. Therefore, we utilize $\gamma$, ranging from 0 to 1, as a coefficient to modify imprecision through spatial information. For a pixel $x$, $\gamma_x$ is calculated in a rectangular window with size $s$ through the ratio of the number of its neighbors sharing the same label, noted as $\nu_x$ and the number of all the pixels in the window, noted as $\nu$:
\begin{equation}
    \gamma_x=\frac{\nu_x}{\nu}
\end{equation}
This coefficient introduces spatial information into mass function, aiming to decrease belief degree for pixels who has different labels than their neighbors.

For supervised learning model, instead of directly choosing training data from satellite image, the spectral model is applied first to generate training samples. Only data with high belief degree of its attributed label can be utilized to train the supervised learning model. Training samples are chosen separately for `water' and `non-water'

Mass function is defined based on the distance to center of class in the feature space. For instance, a pixel $x$, more being away from the center of `water' signifies that it is less credible to make $x$ pertain to this class. Nevertheless, this does not identify falling into `non-water' is more reasonable for $x$, leading to augment in mass value of `ignorance'. Since the supervised model is generated using results from the spectral model, it should be considered to precede the spectral model. This brings these two models into dependent position instead of traditionally independent classifiers in Dempster-Shafer framework. Therefore, the spectral model can regard results from the supervised model as ground truth to calculate its discounting coefficients.

For a pixel $x$, we note $m_2(\{\omega_1\})(x)$ for mass of `water', $m_2(\{\omega_2\})(x)$ for `non-water', $c_{\omega_1}$ for the center of `water' and $c_{\omega_2}$ for the center of `non-water'. $d^2(c_i,x)$ is the distance from $c_i$ to $x$. The mass functions are defined as follows:
$\forall x, d^2(c_{w1},x) \leq d^2(c_{w2},x)$
\begin{equation}
    m_2(\{\omega_1\})(x)=\frac{\alpha}{N}e^{-\frac{d^{2(c_{\omega_1},x)}}{D'_{\omega_1}}}
\end{equation}
\begin{equation}
    m_2(\{\omega_2\})(x)=0
\end{equation}
$\forall x, d^2(c_{w1},x)>d^2(c_{w2},x)$
\begin{equation}
    m_2(\{\omega_1\})(x)=0
\end{equation}
\begin{equation}
    m_2(\{\omega_2\})(x)= \frac{\alpha}{N}e^{-\frac{d^{2(c_{\omega_2},x)}}{D'_{\omega_2}}}
\end{equation}
For the discernment $\Omega$, the mass function is defined as follows:
\begin{equation}
   m_2(\{\Omega\})(x)=1-m_2(\{\omega_1\})(x)-m_2(\{\omega_2\})(x)
\end{equation}
The coefficients $D'_{\omega_1}$ and $D'_{\omega_2}$ are also the normalized coefficient on account of the great distance in the two centers. $D'_{\omega_1}$ is the largest distance from $c_{\omega_1}$ to all pixels who are closer to $c_{\omega_1}$ than $c_{\omega_2}$ while $D'_{\omega_2}$ is defined similarly on the contrary. 
\begin{equation}
    D'_{w1} = \mathrm{max}\ \ d^2(c_{w1},x_{w1})
\end{equation}
\begin{equation}
    D'_{w2} = \mathrm{max}\ \ d^2(c_{w2},x_{w2})
\end{equation}
where $x_{w1}$ refers to the pixels closer to $c_{w1}$ than $c_{w2}$ while $x_{w2}$ is the pixels more approaching to $c_{w2}$ instead of $c_{w1}$.

Since we are in a globally unsupervised context, we directly make discounting coefficient $\alpha$ equal to $0.95$

\section{Fusion between spectral model and supervised model}
\label{Fusion}
As we have explained before, the spectral model and the supervised model are not independent since the former provides training samples to the latter, signifying the supervised model has more reliable results than the spectral model. Therefore, it is necessary to modify mass function of the spectral model in terms of supervised classification before fusion. 

Instead of traditionally measuring reliability of source, two discounting coefficients $\alpha_{\omega_1}$ and $\alpha_{\omega_2}$, used as penalty, are devoted in the spectral model to reduce mass value of pixels having different labels with the supervised model. On the contrary, those who have the same label in the two models keep the initial value of their discounting coefficients. Therefore, $\alpha_{\omega_1}$ and $\alpha_{\omega_2}$ are not parameters controlling the region of 'ignorance'. The coefficients $\alpha_{\omega_1}$ and $\alpha_{\omega_2}$ are calculated through the confusion matrix, considering results from supervised learning as true label, and are givn by:
\begin{equation}
   \alpha_{\omega_1} = p(\theta_{\omega_1}\lvert\vartheta_{\omega_2})
\end{equation}
\begin{equation}
   \alpha_{\omega_2} = p(\theta_{\omega_2}|\vartheta_{\omega_1})
\end{equation}

We note {\em via} $\vartheta_{\omega_1}$ the label of pixel $x$ assigned by supervised model as `water' or and $\vartheta_{\omega_2}$ as `non-water', and $\theta_{\omega_1}$ and $\theta_{\omega_2}$ for the counterparts in the spectral model.  

The two models employed here are not independent for the reason that training samples used in the supervised model was generated from the spectral model. And the discounting coefficient of the spectral model was updated based on the supervised model. Traditional combination rules in Dempster-Shafer framework only deal with the situation in which all the sources are independent. Thus, we applied the average rule of combination which allows to consider multiple perspectives from different dependent sources. The average rule of combination is calculated as follows:
\begin{equation}
    m_{1,2}(\{A\})(x) = \dfrac{1}{2}(m_1(\{A\})(x)+m_2(\{A\})(x))
\end{equation}
where $m_1(\{A\})(x)$ represents mass value for a pixel $x$ belonging to $A$ ($A\in 2^\Omega$) according to the spectral model and $m_2(\{A\})(x)$ is the counterpart from the supervised model.

In order to make decision for both singletons and ignorance, the Appriou's rule is applied in our approach, as illustrated previously in equation~\eqref{ap}. For the coefficients in equation~\eqref{md}, $K_d$ and $\lambda_x$ are equal to $1$ and $r$ was chosen as $0.1$.

\begin{figure}[htb]
\centering
\includegraphics[scale=0.5]{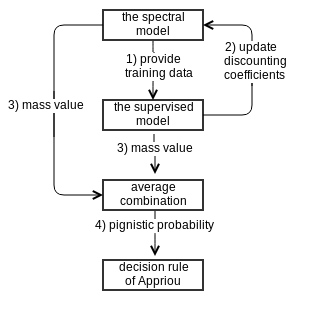}
\caption{principle steps in proposed method}
\label{flow}
\end{figure}

Principle steps of this proposed approach are shown in figure~\ref{flow} with following details:
\begin{enumerate}
    \item Use the spectral model first.
    \item Calculate mass function of `water', `non-water', where discounting coefficients are initialized as 1.
    \item Choose pixels randomly from each class with a relatively high mass value as the training data for supervised model, such as SVM. 
    \item Use the supervised model to predict data.
    \item Calculate mass function of the supervised model.
    \item Update discounting coefficient in the spectral model by results of the supervised model.
    \item Utilize average rule to combine the two classification results.
    \item Calculate pignistic probability.
    \item Apply decision rule of Appriou to attribute labels for pixels.
\end{enumerate}

As explained previously, the fusion is not directly conducted between  both models since one is trained relying on the other. It is crucial to update mass value of the spectral model before fusion step. Since both models are dependent, we choose the average combination rule during fusion. 
The `ignorance' is also considered as a label in final result through the decision rule of Appriou, which presents indistinguishable pixels due to restriction of the model itself and lack of specific spectral information. The results of the proposed method are presented in the next section.
\section{Experiment}
\label{expe}
\begin{figure}[htb]
\centering
\includegraphics[scale=0.3]{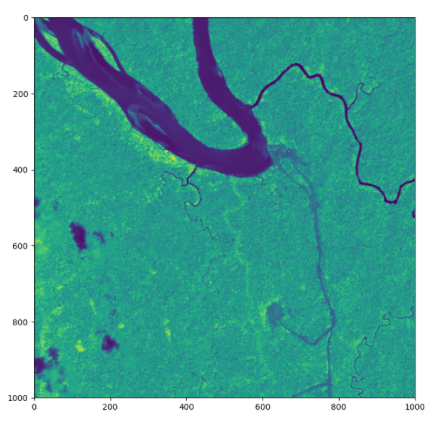}
\label{org}
\caption{Original image in NIR band}
\includegraphics[scale=0.31]{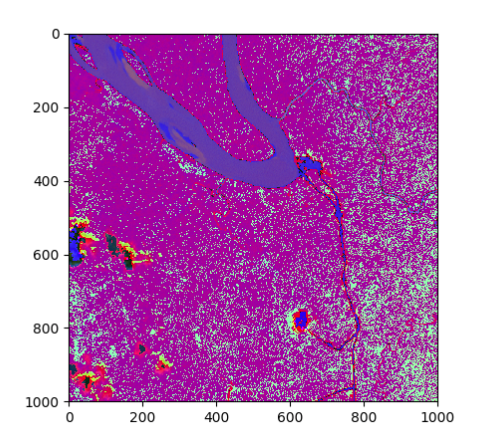}
\caption{Original image in [NDVI, NDWI, RE\_NDWI].}
\label{fs}
\end{figure}

In this section, we compared results from the spectral model, SVM with training samples from the spectral model and result after fusion in the proposed method. Our experiment was conducted in RapidEye data with resolution of 5~m from the study area located in Papua New Guinea, consisting in five distinct bands of the electromagnetic spectrum, as shown in table \ref{tab:re}. For the supervised method, SVM was applied in our study while it is also flexible to choose other supervised learning methods. The original image from RapidEye is shown in figure~\ref{org} and figure \ref{fs}, where NIR band is for the spectral model and the three dimensional space composed of NDVI, NDWI, RE\_NDWI is used for the spectral model. 
\begin{table}[H]
\centering
\caption{Spectral bands of RapidEye }
\label{tab:re}
\scalebox{0.78}{
\begin{tabular}{|c|c|c|}
\hline 
channel & spectral band name & spectral coverage (nm)  \\ 
\hline 
1 & Blue & 440-510 \\ 
\hline 
2 & Green & 520-590 \\ 
\hline 
3 & Red & 630-685 \\ 
\hline 
4 & Red-edge & 690-730 \\ 
\hline 
5 & NIR(Near-Infrared) & 760-850 \\ 
\hline 
 
\end{tabular} }
\end{table}
\vspace{0.1mm}

\begin{figure}[h]
\centering
\includegraphics[scale=0.36]{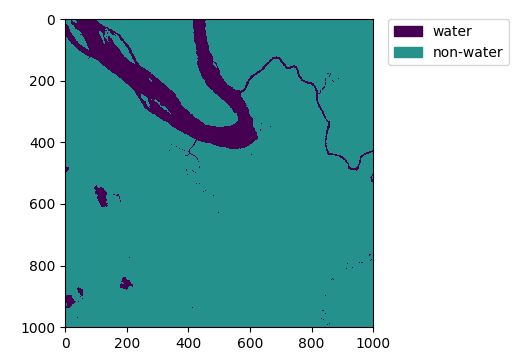}
\caption{Classification results from the spectral model.}
\label{mod}
\vspace{0.1em}
\centering
\includegraphics[scale=0.36]{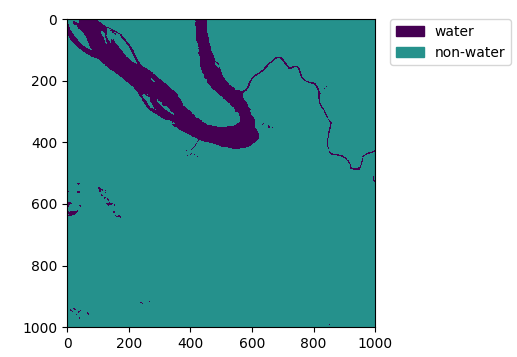}
\caption{Classification results from the supervised model.}
\label{sup}
\end{figure}
The spectral model renders a threshold equal to $5427.18$ in NIR band. For all the data, it labels 8.44\% as `water' and 91.56\% as `non-water'. As shown in \ref{mod}, it identifies the principle water bodies and small branches, also including some parts of thin clouds and shadows. The pixels approaching to the threshold in NIR bands often consist in confusing object, which show similar reflectance in NIR, therefore it is more reasonable to attribute label `water' or `non-water' to the pixels preserving certain distance from the threshold.  

As we explained previously, training samples for SVM was provided by the spectral model. In our study, we took labeled data with a mass value superior to 0.7 from the spectral model because a mass value higher than 0.7 can be considered that the attributed labels are reliable, providing trustful information to the supervised model, and it also guarantees enough training samples as well. This value could be selected flexibly according to the study area and expected accuracy of the approach. 
The supervised model here separates clouds and shadows from surface water owing to high belief degree of training samples, as shown in figure~\ref{sup}. For the data, it classifies 7.64\% as `water', 92.36\% as `non-water'. Apparently, water can be more effectively detected compared to the result from the spectral model, especially in extracting water from very thin clouds. 

\begin{figure}[H]
\centering
\includegraphics[scale=0.3]{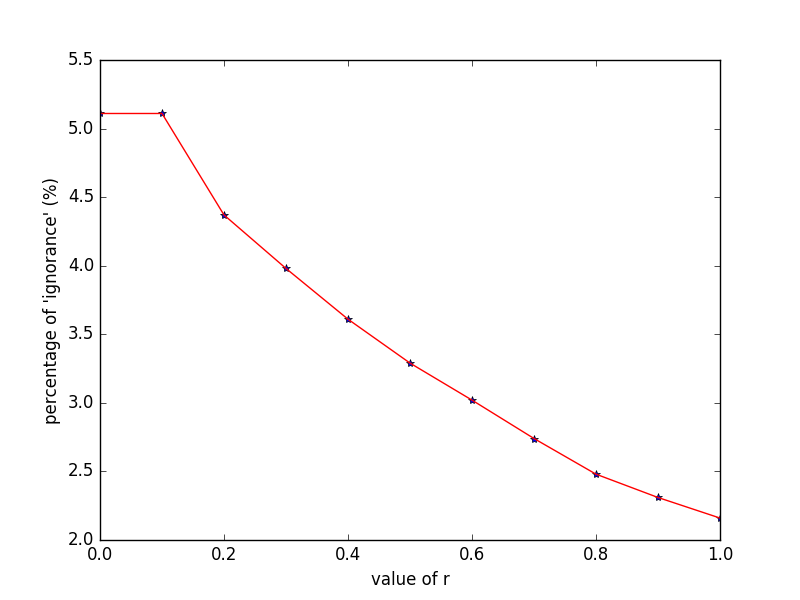}
\caption{The relation between $r$ and percentage of `ignorance'.}
\label{r}
\end{figure}

\begin{figure}[h]
\centering
\includegraphics[scale=0.38]{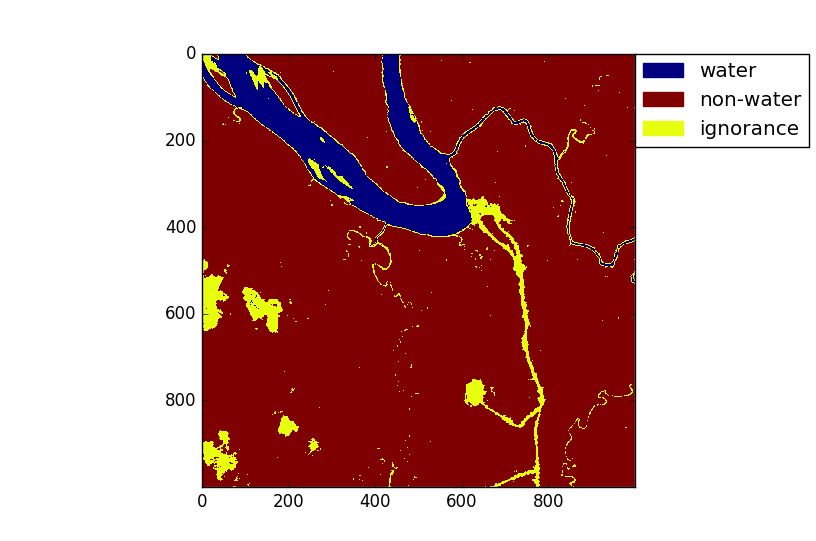}
\caption{The proposed approach with $r=0.1$.}
\label{final0.1}

\centering
\includegraphics[scale=0.38]{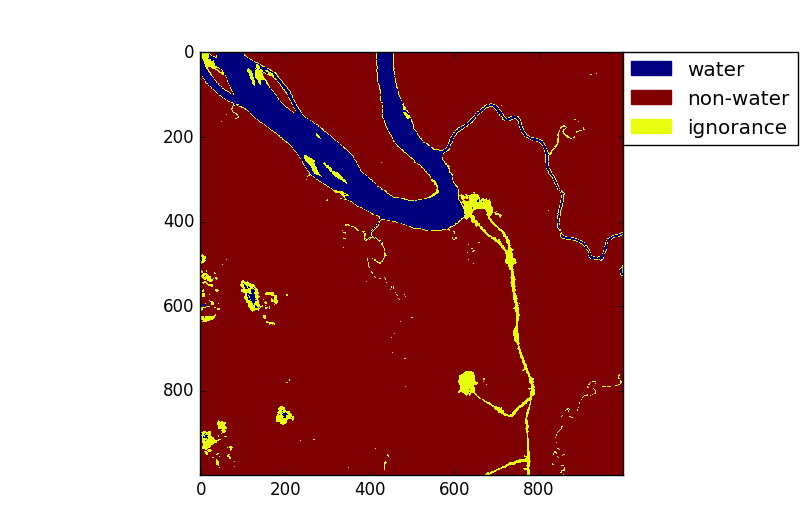}
\caption{The proposed approach with $r=0.5$.}
\label{final0.5}

\centering
\includegraphics[scale=0.38]{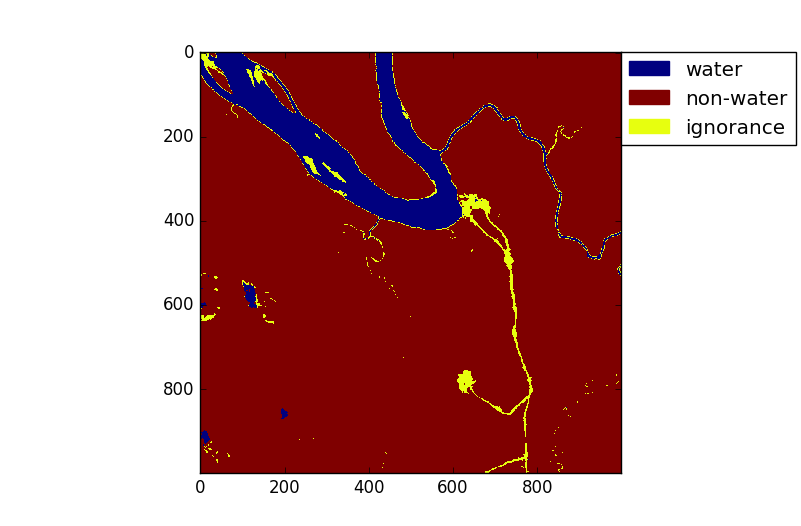}
\caption{The proposed approach with $r=0.9$.}
\label{final0.9}
\end{figure}

As illustrated in section~\ref{DS}, the decision rule of Appriou was applied in our case. In the experiment, $\lambda_X$ in equation~\ref{md} is set to $1$ since we systematically announce the whole frame of discernment when there are only two singletons: `water' and `non-water'. For the parameter $r$ within $[0,1]$, it enables us to make decision ranging from the choice of a singleton to total ignorance, controlling the region of `ignorance'. Various values of $r$ were tested in the experiment and the relation between $r$ and the region of `ignorance' is shown in figure~\ref{r}.
The region of `ignorance' reduces little by little when $r$ changes from $0$ to $1$, showing a nearly linear relation. The larger $r$ is, the less consideration is taken in `ignorance'. Due to the limitation of the space, only results with $r=0.1$, \linebreak $r=0.5$ and $r=0.9$ are shown here in figures~\ref{final0.1}, \ref{final0.5}, \ref{final0.9}.

\begin{figure}[H]
\centering
\includegraphics[scale=0.3]{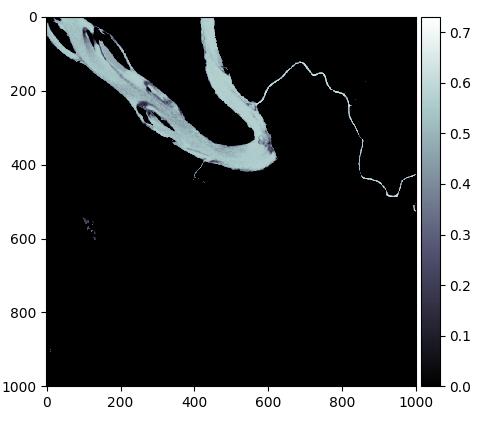}
\caption{Mass values of `water' after fusion.}
\label{abw}

\centering
\includegraphics[scale=0.3]{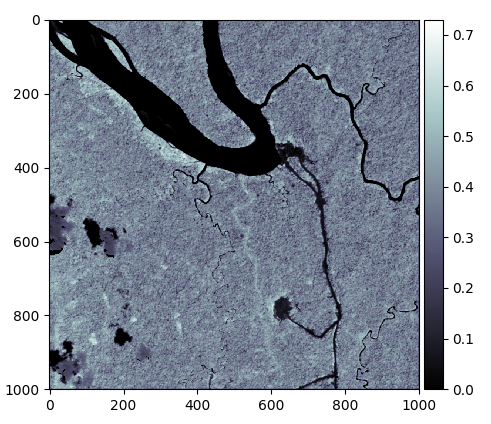}
\caption{Mass values of `non-water' after fusion.}
\label{nw}

\centering
\includegraphics[scale=0.3]{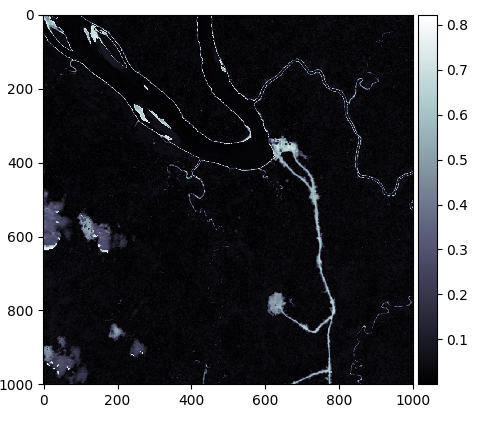}
\caption{Mass values of `ignorance' after fusion.}
\label{pw}
\end{figure}

Compared with both previous results, the proposed approach not only efficiently identify principle river bodies but also show clearly `ignorance' which includes all confusing objects that can not be distinguished directly, providing valuable information for further land cover classification. Comparison between the results and original satellite data indicates the `ignorance' here is composed of thin clouds and its shadows, a route and half-dry watery areas, such as edge of rivers and extremely small and shallow stream-way on the ground. For study data, the proposed method with $r=0.1$ identifies 7.41\% as `water', 87.92\% as `non-water' and 4.67\% as `ignorance'. Mass values after fusion for each class are shown in figures~\ref{abw}, \ref{nw}, and \ref{pw}. Although the distance for confusing object to the NIR threshold in the spectral model is hard to measure, the proposed approach allows to signify all confusing object through `ignorance', clearly improving accuracy of water detection.

Due to the lack of ground truth, we verify our approach through the comparison of the original multi-spectral images and classification results manually through ENVI, a specific software for processing and analyzing geospatial imagery, which allows to show meaningful information from imagery. The proposed approach signifies a satisfying ability to find nearly all the principle rivers bodies and its little branches, separating water bodies from clouds and shadows. For the `non-water' class, the proposed approach also displays very satisfying results in our manual verification in ENVI, in which nearly all the non-water areas were correctly identified.
Furthermore in the class `ignorance', almost all the extremely small and half dry stream-way were detected, also including the objects often confused lightly with water in the land cover classification. The `ignorance' here signifies limitation of the spectral model and the supervised model and also represents insufficiency of spectral information.

\section{Conclusion}
\label{conclusion}
In this article, a new automatic approach in terms of fusion between a spectral model and supervised learning method is proposed, in which Dempster-Shafer evidence theory is applied on dependent sources instead of independent sources. The spectral model and the supervised model is in a serial structure since training samples of the latter was chosen from results of the former, owing to which, discounting coefficients of the spectral model could be calculated considering results of the supervised model as ground truth.

The new approach provides very satisfying performances on detection of water bodies in natural environment. Not only could large water bodies as rivers be detected efficiently, but also small water bodies could be identified from disturbing objects such as clouds. Apparently, the proposed approach overcomes drawbacks of the spectral water as well as unavailability of certain spectral band in our data, such as Shortwave Infrared (SWIR) band. Moreover, `ignorance' gathers all similar objects to water in our classification system, signifying restriction in the two basic models and multi-spectral information themselves. 

In addition, `ignorance' shown in the final results is able to provide valuable information for further land cover classification, especially in a fully automatic and unsupervised context because it helps to specify supplementary information or technology should be applied on its inner objects. We could use supplementary information, for instance, Middle-infrared (MIR) band or some methods specific in identifying built-up or clouds from data labeled as `ignorance' to separate the objects we are interested in.

\begin{IEEEbiography}[{\includegraphics[width=1in,height=1.25in,clip,keepaspectratio]{picture}}]{John Doe}
\blindtext
\end{IEEEbiography}

\bibliographystyle{unsrt}
\bibliography{references}
\end{document}